\documentclass{article}

\usepackage{PRIMEarxiv}

\usepackage{microtype}
\usepackage{graphicx}
\usepackage{booktabs} % for professional tables
\usepackage{xcolor}
\usepackage{physics}
\usepackage{amsfonts}
\usepackage{amssymb}
\usepackage{amsthm}
\usepackage{bbm}
\usepackage[ruled]{algorithm2e}

\usepackage[T1]{fontenc}

\usepackage{longtable}

\usepackage{bm}

\usepackage[square,sort,comma,numbers]{natbib}

\usepackage[T1]{fontenc}
\usepackage[latin9]{inputenc}
\usepackage{color}
\usepackage{calc}
\usepackage{algorithm2e}
\usepackage{amsmath}
\usepackage{amsthm}
\usepackage{amssymb}
\usepackage{graphicx}
\usepackage[final]{pdfpages}
\PassOptionsToPackage{normalem}{ulem}
\usepackage{ulem}

\makeatletter

\makeatletter

\let\oldforeign@language\foreign@language
\DeclareRobustCommand{\foreign@language}[1]{%
  \lowercase{\oldforeign@language{#1}}}
\theoremstyle{plain}

\theoremstyle{plain}

\theoremstyle{definition}

\usepackage[caption=false,font=footnotesize]{subfig}

\theoremstyle{definition}

\theoremstyle{remark}

\numberwithin{equation}{section}

\usepackage{pdflscape}

\usepackage{algorithmic}
\usepackage{amsfonts}
\usepackage{amsmath}

\makeatother

\providecommand{\definitionname}{Definition}
\providecommand{\propositionname}{Proposition}
\providecommand{\theoremname}{Theorem}

\usepackage{amssymb}
\usepackage[figuresright]{rotating}

\usepackage{hyperref}       % hyperlinks
\usepackage{url}            % simple URL typesetting
\usepackage{booktabs}       % professional-quality tables
\usepackage{amsfonts}       % blackboard math symbols
\usepackage{nicefrac}       % compact symbols for 1/2, etc.
\usepackage{microtype}      % microtypography
\usepackage{lipsum}
\usepackage{fancyhdr}       % header
\usepackage{graphicx}       % graphics
\graphicspath{{media/}}     % organize your images and other figures under media/ folder

\usepackage{academicons}
\definecolor{orcidlogocol}{HTML}{A6CE39}
 
\title{Quantum Adaptive Fourier Features for Neural Density Estimation
\thanks{\textit{\underline{Citation}}: 
\textbf{Joseph et al., Quantum Adaptive Fourier Features for Neural Density Estimation.}} 
}

\author{
  Joseph A. Gallego M.$^1$, Fabio A. Gonz\'{a}lez$^2$ \\
  MindLab \\
  Universidad Nacional de Colombia \\
  Bogot\'{a}, Colombia\\
  \texttt{\{jagallegom, fagonzalezo\}@unal.edu.co} 
}

\begin{document}

\maketitle

\begin{abstract}
%% Text of abstract
Density estimation is a fundamental task in statistics and machine learning applications. Kernel density estimation is a powerful tool for non-parametric density estimation in low dimensions; however, its performance is poor in higher dimensions. Moreover, its prediction complexity scale linearly with more training data points. This paper presents a method for neural density estimation that can be seen as a type of kernel density estimation, but without the high prediction computational complexity. The method is based on density matrices, a formalism used in quantum mechanics, and adaptive Fourier features. The method can be trained without optimization, but it could be also integrated with deep learning architectures and trained using gradient descent. Thus, it could be seen as a form of neural density estimation method. The method was evaluated in different synthetic and real datasets, and its performance compared against state-of-the-art neural density estimation methods, obtaining competitive results.   
\end{abstract}

% keywords can be removed

%%
%% Start line numbering here if you want
%%
% \linenumbers

%% main text
\section{Introduction}
\label{sec:introduction}

The estimation of the joint distribution, $p(x_1,\cdots,x_n)$, of a set of random variables is a general important task in machine learning. This estimation of the underlying distribution has a variety of applications, for instance: density estimation, anomaly detection, non-supervised, and supervised learning. Kernel density estimation (KDE) can approximate  arbitrary density functions and its performance increases when more data points are available \citep{rosenblatt1956, parzen1962estimation}. The drawback of this method is that it requires all the training data points to make a prediction, which makes it a memory-based method \citep{Chen2017}. State-of-the-art approximations methods of KDE such as space partitioning \citep{gray2003nonparametric}, random sampling \citep{march2015askit}, and hashing based estimators \citep{Charikar2018, Backurs2019} try to overcome this issue; nonetheless, their prediction complexity increases with more training data points \citep{Siminelakis2019}. The new approach proposed in this paper rely on quantum machine learning and random Fourier features and will be discussed in Section \ref{sec:methods}. This new method is an approximation of kernel density estimation and its prediction complexity is constant with respect to the number of training points.

A different approach to density estimation is based on neural networks with deep architecture. This approach is called neural density estimation and one of its main advantages is that it can be integrated with other deep-learning architectures. The most representative approaches to neural density estimation are autoregressive neural models, normalizing flows and generative adversarial models.

% One of the first attempts was Restricted Boltzman Machines \citep{smolensky1986information, freund1992fast, hinton2002training}. However, this method is based on the calculation of the partition function. In general, this calculation for more than 30 hidden variables is forbidden \citep{larochelle2011neural}. \citet{larochelle2011neural, frey1998graphical, bengio1999modeling} proposed a new kind of models called autoregressive neural models. They use the conditional probability rule to obtain an estimate of the distribution. Normalizing Flow models were proposed in the last decades as an improvement of autoregressive flows models, whose strength is based on the change of variables \citep{dinh2014nice, Rezende2015}. This change of variable can be composed in a series of differentiable and invertible transformations of a known density function, for instance, the normal distribution. The change of variables needs to preserve the volume, which impose a constraint on the availability of base density functions. Nonetheless, these normalizing flow algorithms are really difficult to tune, thus their convergence is not always guaranteed \citep{liu2021density}.

Gonz\'{a}lez et al. \cite{gonzalez2021learning} showed that a combination of random Fourier features and density matrices can be used to perform density estimation. Density matrices are a formalism used in quantum mechanics to represent the state of a quantum system. Its application in machine learning, and in particular in density estimation, has been limited, but that initial exploration suggested that this approach could be a competitive alternative \cite{Gonzalez2020}. In this paper we present a neural density estimation method that approximates the kernel function of KDE using adaptive Fourier features combined with density matrices and systematically evaluate it in different benchmark tasks. One of the main advantages of this approach is that it provides an efficient prediction method whose complexity does not depend on the number of training samples, and also its implementation as a computational graph that is differentiable and thus integrable with deep learning architectures that can be trained by gradient descent. An important characteristic of the model is that, for some applications, it is also possible to train it without using optimization, which alleviates the computational burden associated to gradient-based optimization methods. The method is systematically evaluated on different density estimation tasks and compared against state-of-the-art neural density estimation methods.

The rest of the paper is organized as follows. Section \ref{sec:background} reviews density estimation and related work. Section \ref{sec:methods} presents the novel neural density estimation model based on adaptive Fourier features and density matrices. Section \ref{sec:experimental_evaluation} presents the evaluation of the method in different datasets and its comparison against state-of-the-art neural density estimation methods. Finally, Section \ref{sec:conclusions} presents the conclusions and ideas for future work.

\section{Density Estimation} \label{sec:background}

Given a random variable $X$ and its associated probability distribution $\text{Pr}[X]$, a probability density function is a measurable function $f$ with the property that:
$$
\text{Pr}[X\in A] = \int_A f(x)dx
$$

The problem of density estimation consist on estimating $f$ from a set of \textit{iid} values $\bm{x_1}, \bm{x_2}, \cdots, \bm{x_n}$ sampled from $\text{Pr}[X]$.

Density estimation methods can be broadly divided into parametric and non-parametric approaches. The former assume a parametric model $q_\phi(\bm{x})$ that has adjustable parameters $\phi$, and an optimization process is performed to make $q_\phi(\bm{x})$ as close as possible to the underlying density function $p(x)$. For instance, we can assume that the underlying data comes from a Gaussian (Normal) distribution. However, a drawback is that simple parametric models cannot represent arbitrarily complex density functions. More complex ones, such as mixture models, require several components to approximate density functions compared to compact forms such as kernel density estimation.   

\subsection{Kernel Density Estimation}
 Kernel density estimation was independently proposed by Emanuel Parzen \citep{parzen1962estimation} and Murray Rosenblat \citep{rosenblatt1956} in its present form. Let $\bm{x_1}, \bm{x_2}, \cdots, \bm{x_n}$ be $iid$ training data points drawn from a particular, unknown distribution on $\mathbb{R}^d$. Let $\bm{x}$ be a new sample whose density estimate is desired; we can obtain an estimate of $\bm{f(x)}$ as the mean of a kernel function evaluated on $\bm{x}$ and $\bm{x_i}$, as follows: \\

        \begin{equation}
            \hat{f}(\bm{x})= \frac{1}{M_\gamma n} \sum_{i=1}^n k_\gamma(\bm{x_i},\bm{x})
            \label{eq:kernel_density_estimation}
        \end{equation}
        
where $\gamma$ is the bandwidth parameter, $k_\gamma(\cdot)$ is a kernel function such as the Gaussian kernel and $M_\gamma$ is a normalizing constant depending on the bandwith parameter. This method converges to the true distribution function with more training data points, but its linear complexity for estimate the density of a new $\bm{x}$ data point in terms of the training data points is prohibitive on current large data sets.

\subsection{Neural Density Estimation} \label{subsect:neural_flows}

Three main approaches have been used in state-of-the-art neural density estimation: autoregressive models, normalizing flows and generative adversarial networks. Autoregressive methods have their origin in the restricted Boltzman machine, which is a Markov random field with bipartite substructure, where a connection is established between the weights $\bm{W}$ and the observations $\bm{v}$. One issue of this kind of method is their intractable Z partition function who ensure a valid distribution and sums to 1  \citep{larochelle2011neural, frey1998graphical, bengio1999modeling}. 

Normalizing Flow models were proposed in the last decades as an improvement of autoregressive flows models, whose strength is based on the change of variables \citep{dinh2014nice, Rezende2015}. This change of variable can be composed in a series of differentiable and invertible transformations of a known density function, for instance, the normal distribution. The change of variables needs to preserve the volume, which imposes a constraint on the availability of base density functions. Nonetheless, these normalizing flow algorithms are really difficult to tune, thus their convergence is not always guaranteed \citep{liu2021density}.

In \citep{liu2021density}, the authors propose a new algorithm for density estimation using deep generative neural networks. In the case of the discriminators, the $z$ discriminator is used to distinguish the generated latent variable $\hat{z}$ from the real latent variable $z$. The other discriminator is used to discern the true data $x$ from the generated data $\hat{x}$. This method can use complex deep neural networks as discriminators, e.g., convolutional neural networks or transformers.

The model presented in this paper follows a different approach to neural density estimation which is based on density matrices and kernel-approximating Fourier features. One of the main advantages of this approach is its simplicity as well as its good performance in some benchmark tasks as shown by the experimental evaluation in Section \ref{sec:experimental_evaluation}. In this section different state-of-the-art neural density methods were used as baselines. These methods are described next:

% Normalizing flows models data $\bm{x}$ using a sequence of invertible and differentiable transformation of a generally simple $f$-function. 

% The method uses a change of variables where $Z$ and $X$ are defined as random variables, such that $X = f_{\theta}(Z)$, $Z = f_{\theta}^{-1}(X)$, and $f:\mathbb{R}^n \rightarrow \mathbb{R}^n$. Then,   

% $$p_X(\bm{x};\theta) = p_Z(f_{\theta}^{-1}(\bm{x})) \left| \text{det}\left(\frac{\partial f_{\theta}^{-1}(\bm{x})}{\bm{x}}\right) \right|$$ \label{eq:normalizing-flows}

% The volume is conserved in this transformation due to the calculation of the determinant. Generative adversarial networks is based on two generator networks and two discriminators. One of the generators is used to map from $z$ latent space to $x$ space similar to the process performed by normalization flow algorithms. The other generator is used to perform the opposite transformation from $x$-space to $z$-space. We used the following algorithms as baselines: 

\begin{itemize}
    \item Masked Autoregressive Flow (MAF): use the following recursions for each layer:
$x_i=u_i \exp \alpha_i + \mu_i$ where $\mu_i=f_{\mu_i}(\boldsymbol{u}_{1:i-})$         \citep{Papamakarios2017}. MAF is a generalization of RealNVP. 
    \item Inverse Autoregressive Flow (IAF) \citep{Kingma2016}:
    In \citep{Papamakarios2017}, the authors show that inverse autorregresive flow is a generalization of RealNVP. Define $z_0=(z'_0 - \mu_0)/\sigma_0$ and $z_i=(z'_i-\mu(z'_{1:i-1})/\sigma(z'_{1:i-1})$, then the Jacobian is lower triangular. This implies that the determinant$ |d\bm{z}/d\bm{z'}|$ can be computed as $\prod_{i=1}^D 1/\sigma_i(z_{1:i-1})$ who is not dependent of $z'_i$.

    \item Planar Flow \citep{Rezende2015}: this normalized flow uses a family of transformation of the form $f(\bm{z})=\bm{z} + \bm{u}h(\bm{w}^T\bm{z} + b)$ where $\bm{u}$, $\bm{w}$, $b$ are free parameters and $h(\cdot)$ is a element-wise function. This transformation has a triangular Jacobian.

    \item Real NVP \citep{Dinh2016}: this method uses coupling layers as follows $y_{1:d} = x_{1:d}$ and $y_{d+1:D}=x_{d+1:D} \bigodot exp(s(x_{1:d})) + t(x_{1:d})$ where $s$ and $t$ means scale and translation respectively, and $\bigodot$ is the Hadamard product or element-wise product. Also, the inverse of such transformation do not involve the computation of the inverse of neither $s$ or $t$, therefore these functions can be arbitrarily complex and difficult to invert in particular can be multilayer neural networks.

    \item Neural Spline Flow \citep{Durkan2019}: Neural Splines Flow uses a partition of K *nodes* of the space between (-B,-B) and (B,B). The out-of-range transformation is mapped as the identity. This makes the overall transformation linear out-of-range, so it can take unrestricted inputs. Each knot uses a monotone rational-quadratic function. The authors claim that rational-quadratic functions are easy to derive and, due to their monotonic behavior, are also analytically invertible. 
    \end{itemize}

\section{Quantum Adaptive Fourier Features for Density Estimation (QAFFDE)}
\label{sec:methods}
 
\begin{figure*}[t]
\begin{centering}
\includegraphics[scale=0.40]{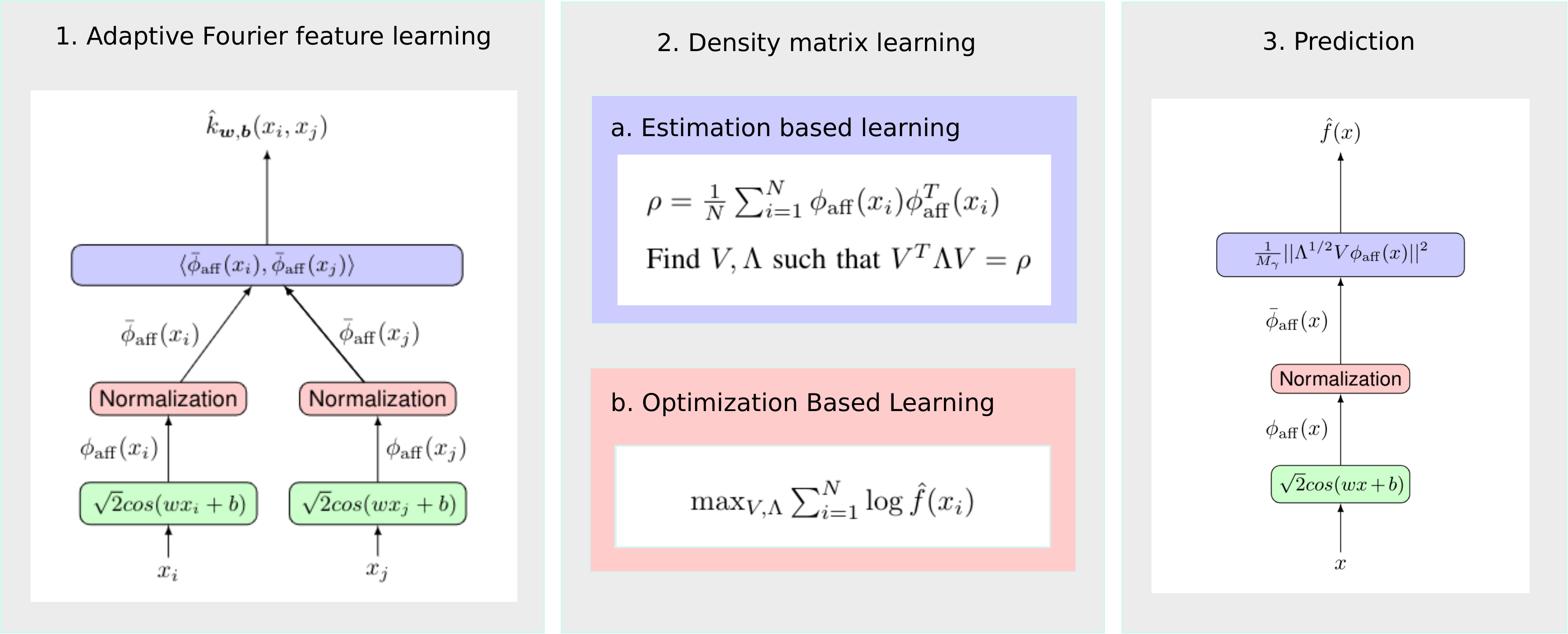}
\par\end{centering}
\caption{Quantum Adaptive Fourier Features for Density Estimation (QAFFDE) method. Step 1, learning of the adaptive Fourier features by a siamese neural network. Step 2, training of the model using an estimation strategy or an optimization strategy to learn the density matrix $\rho$. Step 3, estimation of the density of a new sample using the learn density matrix $\rho=V^T\Lambda V$.   \label{fig:model}}
\end{figure*}
 
In this section, we present a method for neural density estimation that starts from the same idea as kernel density estimation to build a non-memory-based method that is more efficient in prediction and that can be also trained using gradient descent. This method extends the DMKDE method proposed by \cite{gonzalez2021learning}. For accomplishing this we will build an explicit feature map, $\phi:X \rightarrow F$, that approximates the kernel in Equation  \ref{eq:kernel_density_estimation},  $k_\gamma(x,x_i) \approx \langle \phi(\bm{x}), \phi(\bm{x_i})\rangle$. The main components of the method are shown in Figure \ref{fig:model}. Each component is explained in the following subsections.

% \begin{align} \label{eq:kde_in_feature_space}
% \begin{aligned}
% \hat{f}_\gamma(\bm{x}) & = \frac{1}{N}\sum_{i=1}^N k_\gamma(\bm{x}, \bm{x_i})  \\
% & \simeq \frac{1}{N} \sum_{i=1}^N \langle \phi(\bm{x}), \phi(\bm{x_i})\rangle  \\
% & = \langle \phi(\bm{x}), \frac{1}{N} \sum_{i=1}^N \phi(\bm{x_i})\rangle  \\
% \end{aligned}
% \end{align}

% Nonetheless, the explicit map used by Equation \ref{eq:kde_in_feature_space} is not generally available, for instance, the Gaussian kernel is related to a feature space that is an infinite-dimensional Hilbert space . To solve this issue, we used the technique developed by \citet{rahimi2007random} called the random Fourier features, which can approximate the feature space. A more detailed explanation is given in the next subsection.

\subsection{Adaptive Fourier feature learning}

Kernel methods are the backbone of several machine learning algorithms, such as support vector machines \citep{hearst1998support}, Gaussian processes \citep{rasmussen2003gaussian}, kernel density estimation \citep{parzen1962estimation,rosenblatt1956}, kernel principal component analysis \citep{scholkopf1997kernel}, among others. A kernel calculates the dot product in an implicit feature space. This feature space is usually high-dimensional or even of infinite dimension, as it is the case for the Gaussian kernel \citep{Scholkopf2002}. Random Fourier features (RFF) \citep{rahimi2007random} is a method that given a shift-invariant kernel, $k:X\times X \rightarrow \mathbb{R}$, calculates an explicit feature map $\phi_{\text{rff}}:X \rightarrow F$ such that $k(x,y)\approx \langle{\phi}_\mathrm{rff}(x),{\phi}_\mathrm{rff}(x)\rangle$. RFF are based on the Bochner's theorem 
\citep{rahimi2007random} and approximates the kernel by estimating an expected value $
k(\bm{x},\bm{y}) \simeq \mathbb{E}_{\bm{w}}[Z_{\bm{w}}(\bm{x})Z_{\bm{w}}(\bm{y})]
$
where $Z_{\bm{w}}(\bm{x})=\sqrt{2}cos(\bm{w}^*\bm{x} + \bm{b})$, with $\bm{w} \sim \mathcal{N}(0,1)$ and $\bm{b} \sim Uniform[0,2\pi]$ for the Gaussian kernel. The features correspond to a set $\phi_{\mathrm{rff},i}\}_{i=1\dots D}$ with $\phi_{\mathrm{rff},i}(x)=\sqrt{2}cos(\bm{w}_i^*\bm{x} + \bm{b}_i)$ where $\bm{w}_i$ and $\bm{b}_i$ are sampled from the aforementioned distributions. The higher the number of features, the better the approximation.

An interesting characteristic of RFF is that they are data independent.  However it is possible to achieve a better approximation of the kernel with the same number of features if we use data to learn the features instead of the data-agnostic sampling procedure of the original RFF method. Some works have proposed data-dependent strategies to obtain better features: leverage score sampling \citep{Li2019, liu2020random}, reweighted random features \citep{Sinha2016, avron2016quasi}, and kernel learning \citep{Li2019, Bullins2017}. 

In this work we propose a new method to learn the $\bm{w}$ and $\bm{b}$ vectors using gradient descent. We called this approach  adaptive Fourier features (AFF). The method uses a siamese neural network which is shown in the first step of Figure \ref{fig:model}. The neural network is trained by sampling pairs of samples $x_i$ and $x_j$ from the data set and minimizing a square error loss function $L = \left(k(x_i,x_j)-\hat{k}_{\bm{w}, \bm{b}}(x_i,x_j)\right)^2$, as shown in Algorithm \ref{alg:kde-random-fourier-features}.

\begin{figure}[t]
\begin{centering}
\includegraphics[scale=1]{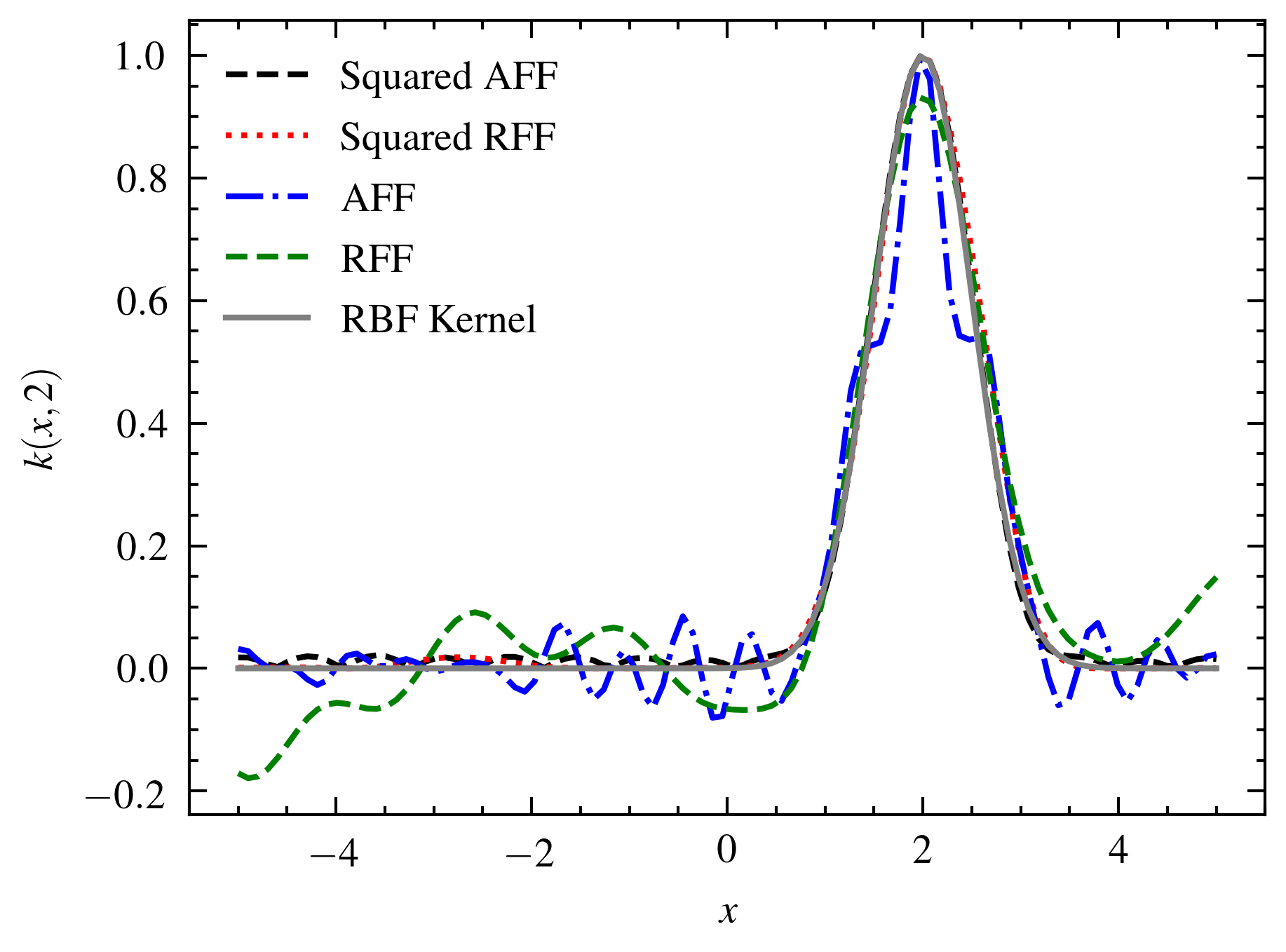}
\includegraphics[scale=1]{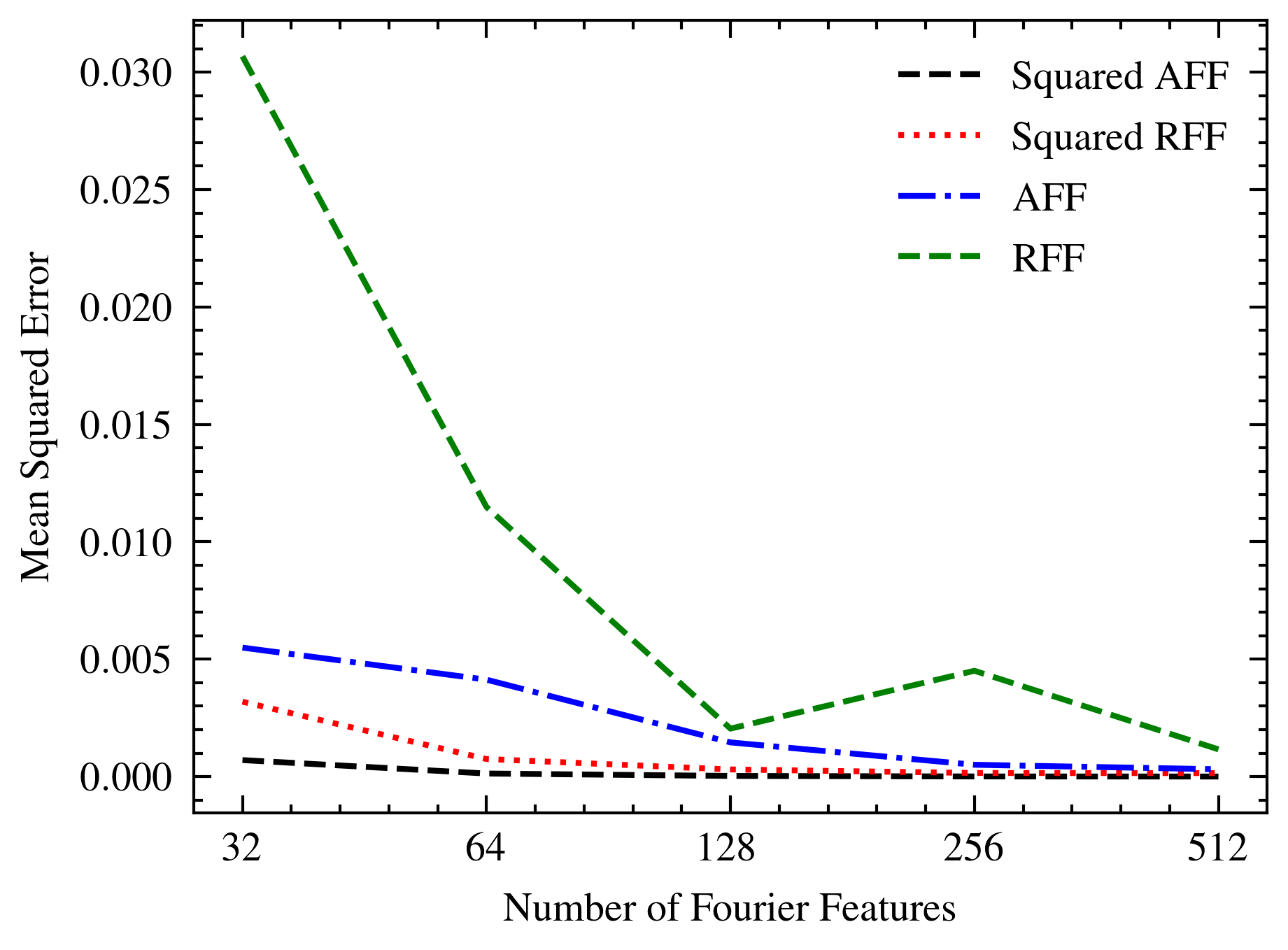}
\par\end{centering}
\caption{(left) Comparison between the real Gaussian kernel centered on 2 as $k(x,y)=\exp{-\gamma ||x-y||^2}$, the random Fourier feature, the random Fourier feature squared, the adaptive Fourier feature, and the adaptive Fourier feature squared. (Right) Mean squared error between approximation Fourier features and the real Gaussian kernel. \label{fig:aff-comparison}}
\end{figure}

 Gonz\'{a}lez et al. \cite{gonzalez2021learning} propose to use the square of the dot product of samples represented using RFF as a better approximation of the Gaussian kernel. We follow the same approach here. Figure \ref{fig:aff-comparison} shows the comparison between the real Gaussian kernel, its approximation using RFF, squared RFF, adaptive Fourier features, and squared adaptive Fourier features . The best approximation is obtained by squared AFF, followed by AFF, squared RFF and RFF. As shown by the right plot in Figure \ref{fig:aff-comparison}, squared AFF can reach a good approximation even with a small number of features, this has a positive impact in the efficiency of the density estimation algorithms presented in the next sections.

\subsection{Kernel Density Estimation using Adaptive Random Features}

If we start from Eq. \ref{eq:kernel_density_estimation}, with $k_{\gamma}$ representing the Gaussian kernel, we can do the following derivation \citep{gonzalez2021learning}:

\begin{align} \label{eq:kde_density_matrix}
\begin{aligned}
\hat{f}(\bm{x}) & = \frac{1}{M_\gamma N}\sum_{i=1}^N k_{\gamma}(\bm{x}, \bm{x_i})  \\
&  = \frac{1}{M_\gamma N}\sum_{i=1}^N k^2_{\gamma/2}(\bm{x}, \bm{x_i})  \\
& \simeq  \frac{1}{M_\gamma N}\sum_{i=1}^N  \langle \phi_\text{aff}(\bm{x}), \phi_\text{aff}(\bm{x_i})\rangle^2 \\
& =  \frac{1}{M_\gamma N}\sum_{i=1}^N  \phi^T_\text{aff}(\bm{x}) \phi_\text{aff}(\bm{x_i})\phi^T_\text{aff}(\bm{x_i})\phi_\text{aff}(\bm{x}) \\
& =  \frac{1}{M_\gamma} \phi^T_\text{aff}(\bm{x}) \left( \frac{1}{N}\sum_{i=1}^N \phi_\text{aff}(\bm{x_i})\phi^T_\text{aff}(\bm{x_i}) \right)
\phi_\text{aff}(\bm{x}) \\
 & = \frac{1}{M_\gamma} \phi(\bm{x})_\text{aff}^T \; \rho \: \phi(\bm{x})_\text{aff}
\end{aligned}
\end{align}

The matrix $\rho$ is called a density matrix \citep{gonzalez2021learning} and it can be seem as a summary of the training data set that can be used to estimate the density of a new sample. Density matrices are a formalism used in quantum mechanics to represent the state of a quantum system. The last line in Eq. \ref{eq:kde_density_matrix} can be seen as an instance of the Born rule that calculates the probability of obtaining a particular state $\phi(\bm{x})_\text{aff}$ when measuring a quantum system whose state is described by the density matrix $\rho$ \citep{gonzalez2021learning}. 

Using directly Eq. \ref{eq:kde_density_matrix} to do estimation of the probability of a new sample could be very inefficient, since the size of training density matrix is $O(D^2)$, where $D$ is the dimension of the AFF. To alleviate this, we can perform a low-rank factorization of $\rho$ as follows:

\begin{equation}
    \rho \approx V^T \Lambda V
\end{equation}

where $V \in \mathbb{R}^{r \times D}$ contains as rows the eigenvectors of $\rho$ corresponding to the $r$ largest eigenvalues, and $\Lambda \in \mathbb{R}^{r \times r}$ is a diagonal matrix with the $r$ largest eigenvalues. In this way, the estimation in Eq. \ref{eq:kde_density_matrix} can be calculated as:

\begin{align} \label{eq:AFDM_estimation}
\begin{aligned}
\hat{f}_\gamma(\bm{x}) & \simeq \frac{1}{M_\gamma} \phi(\bm{x})_\text{aff}^T \rho \phi(\bm{x})_\text{aff} \\
& \simeq \frac{1}{M_\gamma}|| \Lambda^{1/2}V\phi(\bm{x})_\text{aff}||^2
\end{aligned}
\end{align}

Eq. \ref{eq:AFDM_estimation} is the basis for the neural density estimation model presented in the step 3 of Figure \ref{fig:aff-comparison}. The parameters of this model are the weights $W$ and $b$ of the AFF mapping and the $V$ and $\Lambda$ parameters of the density estimation step. The AFF parameters are learned independently by training the neural architecture in step 1 of Figure \ref{fig:aff-comparison} using Algorithm \ref{alg:kde-random-fourier-features}. The parameters $V$ and $\Lambda$ can be learned using two different approaches depicted in Figure \ref{fig:aff-comparison} and detailed in Algorithm \ref{alg:compute_QAFFDE} and \ref{alg:compute_QAFFDE_eigen}. 

The first approach, Algorithm \ref{alg:compute_QAFFDE}, estimates the density matrix $\rho$ from training data and calculates the factorization components $V$ and $\Lambda$ using a spectral decomposition. An important feature of this approach is that it does not requires any optimization, just averaging of the density matrices representing the training samples.

The second approach, Algorithm \ref{alg:compute_QAFFDE_eigen}, exploits the fact that the prediction model (step 3 in Figure \ref{fig:aff-comparison}) is in fact differentiable neural network that can be trained by backpropagation and gradient descent, as it is the widespread practice for neural models. This process is in general more computational demanding than the optimization-less approach of Algorithm \ref{alg:compute_QAFFDE}. Its main advantage is that it can be integrated with other deep architectures and trained jointly. This approach is explored in Section \ref{sect:cond_density_estimation} of the experimental evaluation.

\begin{center}
\begin{algorithm}[t]
 \SetAlgoLined
    \KwIn{Training data set $D=\{x_i\}_{i=1,\cdots,N}$, $\gamma$ kernel bandwidth}
    \KwOut{$w, b$ AFF parameters}
    {Build a set $s=\{(x_i' , x_i'')\}_{i=1,\cdots,m}$ where $x_i'$ and $x_i''$ are randomly sampled from $D$} \\
    {Apply gradient descent to find} \\
    {\quad $w^*, b^* = \arg \min_{w,b} \frac{1}{m}\sum_{x_i,x_j\in s} (k_\gamma(x_i,x_j) - \hat{k}_{w,b}(x_i,x_j))^2$} \\
    \Return $w^*, b^*$
    
    \caption{Adaptive Fourier Feature learning}
    \label{alg:kde-random-fourier-features}
\end{algorithm}
\end{center}
 
\begin{center}
\begin{algorithm}[t]
    
 \SetAlgoLined
    \KwIn{Training dataset $D=\{x_i\}_{i=1,\cdots,N}, w,b$ AFF parameters}
    \KwOut{$V, \Lambda$}
    {Calculate $\rho= \frac{1}{N}\sum_{i=1}^N \phi_\text{aff}(\bm{x_i})\phi^T_\text{aff}(\bm{x_i})$}\\
    {\quad where $\phi_{\text{aff}}(x_i)= \sqrt{2}cos(wx_i + b)$}\\
    {Perform a spectral decomposition of $\rho$}\\
    {\quad $\rho \approx V^T \Lambda V$}\\
     \Return $V, \Lambda$
    \caption{Adaptive Fourier density matrix density estimation training (QAFFDE)}
    \label{alg:compute_QAFFDE}
\end{algorithm}
\end{center}

\begin{center}
\begin{algorithm}[t]
 \SetAlgoLined
    \KwIn{Training dataset $D=\{x_i\}_{i=1,\cdots,N}, w,b$ AFF parameters}
    \KwOut{$V^*, \Lambda^*$}
    {Apply gradient descent to find:}\\
    {\quad $V^*, \Lambda^* = \arg \min_{V, \Lambda} \sum_{i=1}^N \log \hat{f}(x_i)$}\\
    {\quad where $\hat{f}(x_i) =\frac{1}{M_\gamma}|| \Lambda^{1/2}V\phi_\text{aff}(x)||^2$ }\\
     {\quad and $\phi_{\text{aff}}(x_i)= \sqrt{2}cos(wx_i + b)$}\\
     \Return $V^*, \Lambda^*$
    \caption{Adaptive Fourier density matrix density estimation training using gradient descent (QAFFDE-SGD)}
    \label{alg:compute_QAFFDE_eigen}
\end{algorithm}
\end{center}

\section{Experimental Evaluation}
\label{sec:experimental_evaluation}

\subsection{Unconditional Density Estimation} 
\label{subsect:unconditional_density_estimation}
In this subsection we proposed an experiment based on synthetically generated data to evaluate the performance of various neural density estimation methods and compare them with QAFFDE and QAFFDE-SGD. It should be noted that the underlying density function is known for each data set used in the current experiment.

\subsubsection{Data sets and experimental setup}

\begin{figure}[]
\begin{centering}
\includegraphics[scale=0.36]{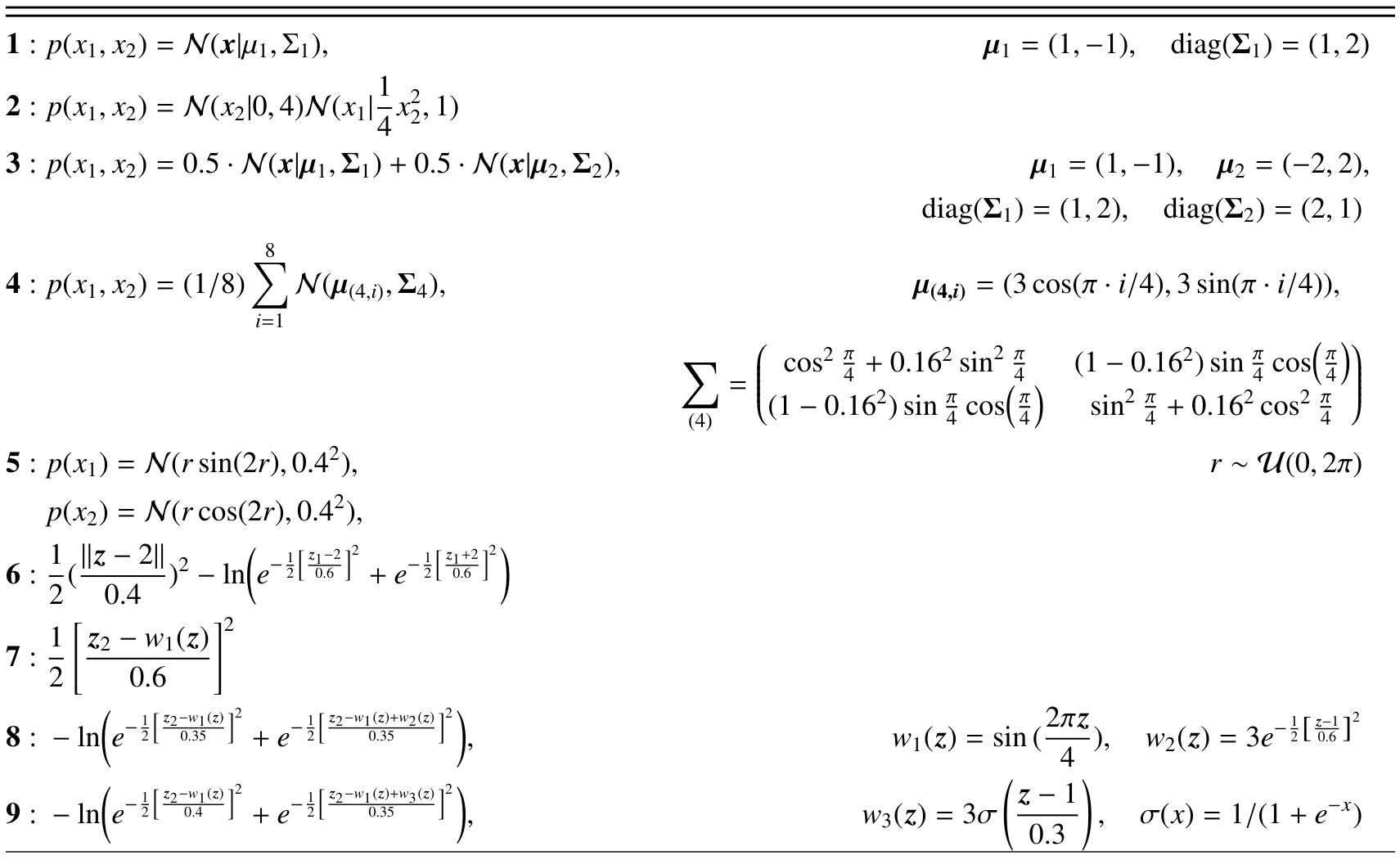}
\par\end{centering}
\caption{Synthetic data set}\label{table:synthetic-data-set}
\end{figure}

Nine synthetic data sets were generated using distinct underlying density functions. Table \ref{table:synthetic-data-set} shows the density estimation function of each synthetic where each number represent the following distributions: (1). Multivariate normal distribution, (2). Arc distribution \citep{Papamakarios2017}, (3). Mixture Gaussian distribution, (4). Star \citep{liu2021density}, (5). Swiss Roll \citep{liu2021density}, (6). Potential 1 , (7). Potential 2, (8). Potential 3, and (9). Potential 4 \citep{Rezende2015}. The symbols used in Table \ref{table:synthetic-data-set} were as follows: . For each dataset we generated 100 000 training data points and 50 000 testing data points. Besides, we assessed four baseline neural flow algorithms: (1). Masked Autoregressive Flow for Density Estimation \citep{Papamakarios2017}, (2). Inverse Autoregressive Flow \citep{Kingma2016} and (3). Planar Flow \citep{Rezende2015}.

All the algorithms were trained using Adam Optimizer with a polynomial decay. We used a cross validation setup along with random search for hyper parameters selection. For each data set and algorithm we tested 30 different combinations of hyper parameters.  For the validation phase, we used 40 000 training data points and 1 000 validation data points. The batch size was set to 64. The initial learning rate for the polynomial decay was searched in the interval $[10^{-5},10^{-1}]$; the final learning rate was set to $10^{-5}$. For each neural flow algorithm, the following setup was used: hidden shape was searched between $20$ and $1000$; the number of layers was searched in the list $2 ,\cdots 24$. For Neural Splines: the number of bins was searched between three and eleven; the $b$-interval was searched between three and seven for each dimension. For QAFFDE: AFF dimension was explored between 250 and 2 000, sigma was explored between $2^{-20}$ and $2^{20}$, and the training of the AFF was performed using $10000$ pairs of different points. In the case of QAFFDE-SGD, we selected the number of eigen-components in the list 0, 0.1, 0.5, 1 where each number represents a percentage of the number of AFF. Moreover, we assessed random initialization and QAFFDE initialization for the density matrix of QAFFDE-SGD.

We used two metrics to evaluate the performance of each algorithm. Those metrics are the mean average error (MAE) and the Spearman's rank coefficient. For each metric, we compute the metrics using the estimate of each algorithm and the real probability. MAE is computed as:  $1/n \cdot \sum_i^n|p(\bm{x_i}) - \hat{p}(\bm{x_i})|$, where $n$ is the number of data points, $p(\bm{x_i})$ is the real probability density, $\hat{p}(\bm{x_i})$ is the estimated probability density given by the method, and $|\cdot|$ is the absolute value function.

\subsubsection{Results and discussion}

\begin{table*}[tbh]
\caption{Spearman correlation for Unconditional Density Estimation Experiment.}
\label{Table:spearman_correlation_unconditional}
\vskip 0.15in
\begin{center}
\begin{sc}
\begin{tabular}{lccccccr}
\toprule
\toprule
        Dataset & QAFFDE & QAFFDE-SGD & MADE & Inverse Maf& Planar Flow & Neural Splines  \\ 
        \toprule
        arc & 0.9943 & 0.9773 & 0.9993 & $\bm{0.9978}$ & 0.7395 & 0.9650 \\ 
        bimodal\_l & $\bm{0.9954}$ & 0.9941 & 0.9924 & 0.9918 & 0.9443 & 0.9892 \\ 
        binomial & $\bm{0.9991}$ & 0.9986 & 0.9991 & $\bm{0.9991}$ & 0.9866 & 0.9988 \\ 
        potential\_1 & $\bm{0.9902}$ & 0.9839 & 0.9516 & 0.9667 & 0.8637 & 0.9723 \\ 
        potential\_2 & $\bm{0.9072}$ & 0.8112 & 0.8478 & 0.8374 & 0.5540 & 0.8175 \\ 
        potential\_3 & $\bm{0.8665}$  & 0.8558 & 0.8059 & 0.8113 & 0.6280 & 0.8125 \\ 
        potential\_4 & 0.8928 & $\bm{0.9094}$  & 0.8768 & 0.8791 & 0.6034 & 0.8816 \\ 
        star\_eight & $\bm{0.9871}$  & 0.9348 & 0.7710 & 0.7349 & 0.5940 & 0.9042 \\ 
        swiss\_roll & $\bm{0.9936}$  & 0.9686 & 0.9713 & 0.9475 & 0.7753 & 0.9451 \\ 
\bottomrule
\end{tabular}
\end{sc}
\end{center}
\vskip -0.1in
\end{table*}

\begin{table*}[tbh]
\caption{Mean average error for Unconditional Density Estimation Experiment.}
\label{Table:mae_unconditional}
\vskip 0.15in
\begin{center}
\begin{sc}
\begin{tabular}{lccccccr}
\toprule
\toprule
        Dataset & QAFFDE & QAFFDE-SGD & MADE & Inverse Maf& Planar Flow & Neural Splines  \\ 
        \toprule
        arc & 0.0052 & 0.0179 & $\bm{0.0004}$  & 0.0007 & 0.0197 & 0.0057 \\ 
        bimodal\_l & 0.0014 & 0.0152 & $\bm{0.0009}$  & 0.0010 & 0.0029 & 0.0012 \\
        binomial & 0.0286 & 0.0266 & $\bm{0.0008}$  & $\bm{0.0008}$  & 0.0032 & 0.0010 \\ 
        potential\_1 & 0.0089 & 0.0735 & 0.0104 & 0.0097 & 0.0235 & $\bm{0.0069}$  \\ 
        potential\_2 & $\bm{0.0273}$  & 0.0548 & 0.0475 & 0.0514 & 0.0619 & 0.0520 \\ 
        potential\_3 & 0.0321 & 0.0231 & 0.0225 & $\bm{0.0218}$  & 0.0340 & 0.0219 \\ 
        potential\_4 & 0.0481 & 0.0377 & 0.0312 & 0.0331 & 0.0451 & $\bm{0.0308}$  \\ 
        star\_eight & $\bm{0.0102}$  & 0.0161 & 0.0196 & 0.0232 & 0.0388 & 0.0172 \\ 
        swiss\_roll & $\bm{0.0028}$  & 0.0344 & 0.0040 & 0.0045 & 0.0223 & 0.0051 \\ 
\bottomrule
\end{tabular}
\end{sc}
\end{center}
\vskip -0.1in
\end{table*}

Table \ref{Table:spearman_correlation_unconditional} shows the results obtained by each of the six neural density estimation algorithms on the 9 synthetic data sets. The results on the ARC, Bimodal and Binomial data sets are similar among the algorithms, excluding Planar Flow whose performance is inferior. The performance of \textbf{QAFFDE} and \textbf{QAFFDE-SGD} is better when compared to the other methods at Potentials 1 to 4, where \textbf{QAFFDE-SGD} is better than \textbf{QAFFDE} at potential 4, which is a really difficult distribution function. The performance of \textbf{QAFFDE} in Star Eight and Swiss Roll is superior compared to the other algorithms. Planar Flow is the worst algorithm among them. Neural Splines has consistent satisfactory results. 

Table \ref{Table:mae_unconditional} shows the results of the mean average error between the actual distribution function and the estimate of each density estimation function on the 9 synthetic data sets. Made has the best MAE in arc, bimodal and binomial.  QAFFDE is the best in potential 2, start eight and swiss roll. Neural splines is the best in potential 1 and 4. QAFFDE-SGD is similar to QAFFDE in terms of MAE results, and its results are close to the best results in Potential 1 to 4, Start Eight and Swiss Roll.

\textbf{QAFFDE} is not the best one in terms of MAE. However, it is worth noting that for the majority of real applications such as classification,  anomaly detection, and regression, the real value of the density function is not really important. The most important property is being able to differentiate between low density areas and high-density areas. This property is intrinsically captured by Spearman correlation.

\begin{figure*}[tbh]
\begin{centering}
\includegraphics[scale=0.17]{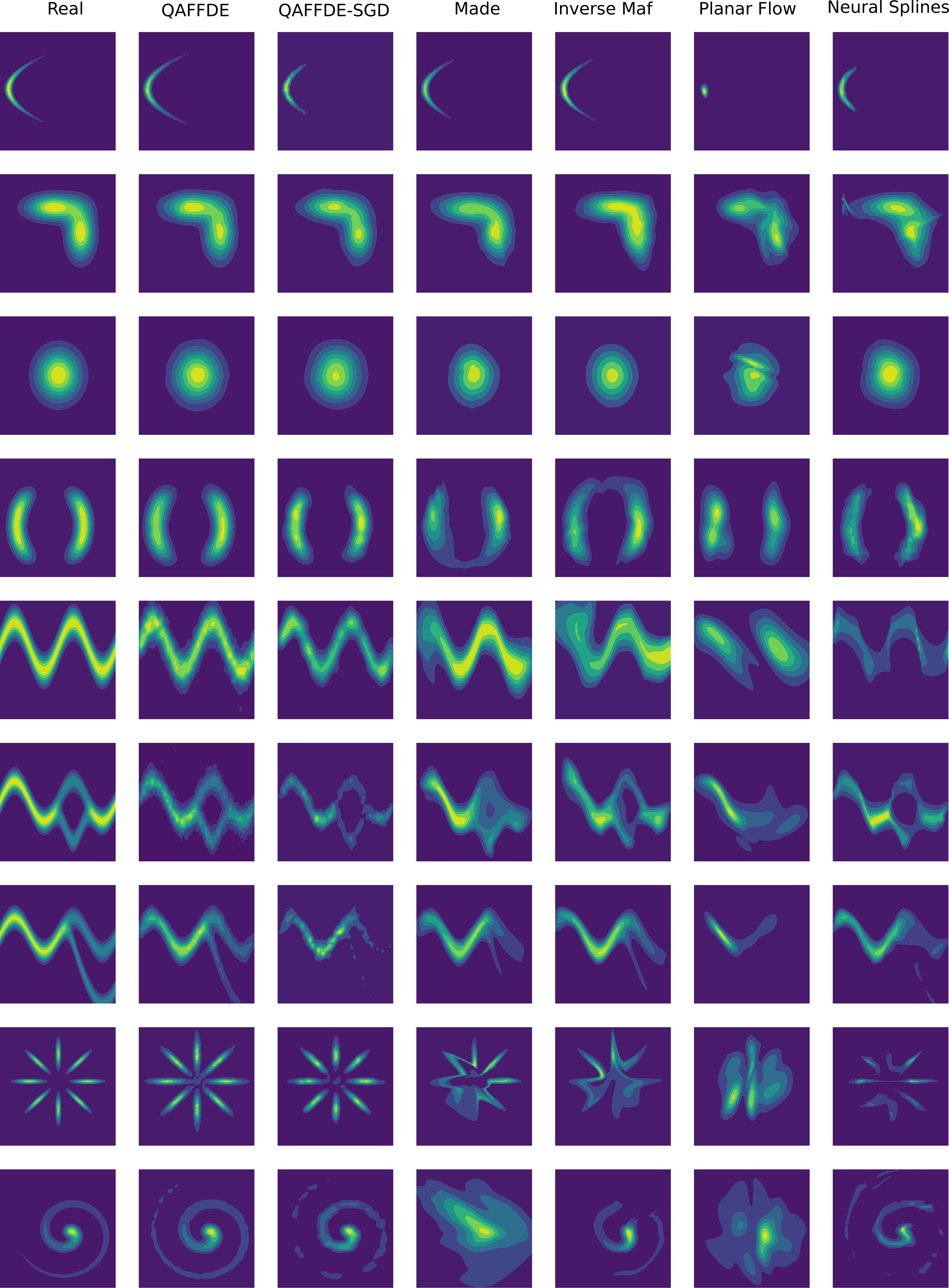}
\par\end{centering}
\caption{Each graph represents the density estimate obtained by a specific algorithm on a two-dimensional data set. From left to right the algorithms are Real Density, QAFFDE, QAFFDE-SGD, Made, Inverse Maf, Planar Flow, and Neural Splines. From top to bottom the data sets are Arc, Bimodal Gaussian, two-dimensional Gaussian distribution, Potential 1-4, Star, and Swizz Roll.\label{fig:mesh-density}}
\end{figure*}

\begin{figure*}[tbh]
\begin{centering}
\includegraphics[scale=0.16]{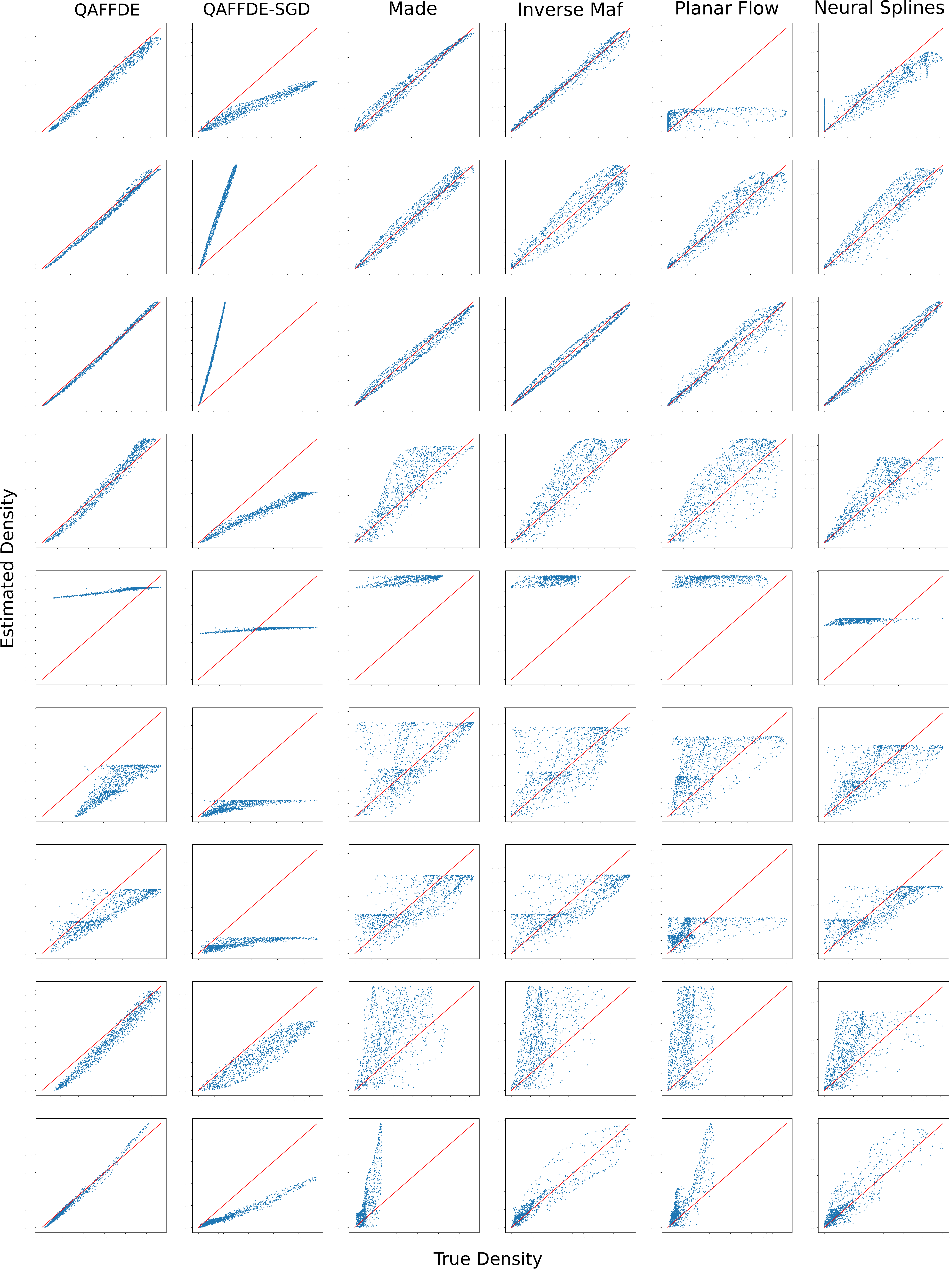}
\par\end{centering}
\caption{Each graph represents the comparison of the Spearman's correlation between the real density for each data set and the density estimate obtained by each algorithm. From left to right the algorithms are QAFFDE, QAFFDE-SGD, Made, Inverse Maf, Planar Flow, and Neural Splines. From top to bottom the data sets are Arc, Bimodal Gaussian, 2-dimensional Gaussian distribution, Moons, Potential 1-4, Star, and Swizz Roll.\label{fig:spearmans-correlation}}
\end{figure*}

Figure \ref{fig:mesh-density} shows the density estimate obtained by applying the six algorithms on the nine two-dimensional synthetic data sets.
The first column is \textbf{QAFFDE}. Its results are the best amongst the six algorithms. It was able to regenerate every data set including the most difficult ones such as Potential 3 and Potential 4. The second columns is \textbf{QAFFDE Sgd}. Its results are similar to those of \textbf{QAFFDE} in seven of nine datasets. It obtained a different results in Potential 4 and Swizz Roll. Figure \ref{fig:spearmans-correlation} shows the comparison between the real density and the density estimate obtained on the nine synthetic data sets when applying the six different algorithms. \textbf{QAFFDE} (first column) shows good results and lower scatter in all data sets. In the first and the second data set, it shows slightly more dispersion than Made. In the Swizz Roll it shows the best performance of all the algorithms. The \textbf{QAFFDE-SGD} algorithm (second column) shows a tendency to over estimate points of low density points given presumably by its log likelihood optimization approach.

\subsection{Unconditional Random Density Estimation in Higher Dimensions}
Classical methods for density estimation, such as Gaussian Mixture Models \citep{li1999mixture} or Kernel Density Estimation \citep{parzen1962estimation, rosenblatt1956}, suffer with higher dimensions due to the curse of dimensionality. Several papers claim that normalizing flow methods can solve this problem. Related to QAFFDE, it potentially can inherit the problems from kernel density estimation; however, as shown in the following experiment, it can deal with higher dimensions and obtain good density estimates. In this subsection, we propose an experiment to show the robustness of QAFFDE methods when the number of dimensions varies and its increased.

\subsubsection{Data sets and experimental setup}

In this experiment, we generate data points from a mixture of random independent Gaussian distributions. For n dimensions, we generated $10 \cdot n$ Gaussian distributions for $n \in [1, \cdots, 10]$. The $\mu_i$ parameters of the Gaussians were generated from a uniform distribution $\mu_i \in (0,1)^n$. The covariance matrices, $\Sigma_i$, were generated with a vector of uniform values as eigenvalues. With this vector, we compute the algorithm proposed by Davies et al. \cite{davies2000numerically} to generate a random correlation matrix. Using both, the centroid and the covariance matrix, we sampled equal number of points from each Gaussian distributions to obtain 40 000 training data points and 10 000 testing data points. Figure \ref{fig:spatial-gmm-results} shows an example in two dimensions of a the random generated samples.

\subsubsection{Results and discussion}
\begin{figure}[t]
\begin{centering}
\includegraphics[scale=0.48]{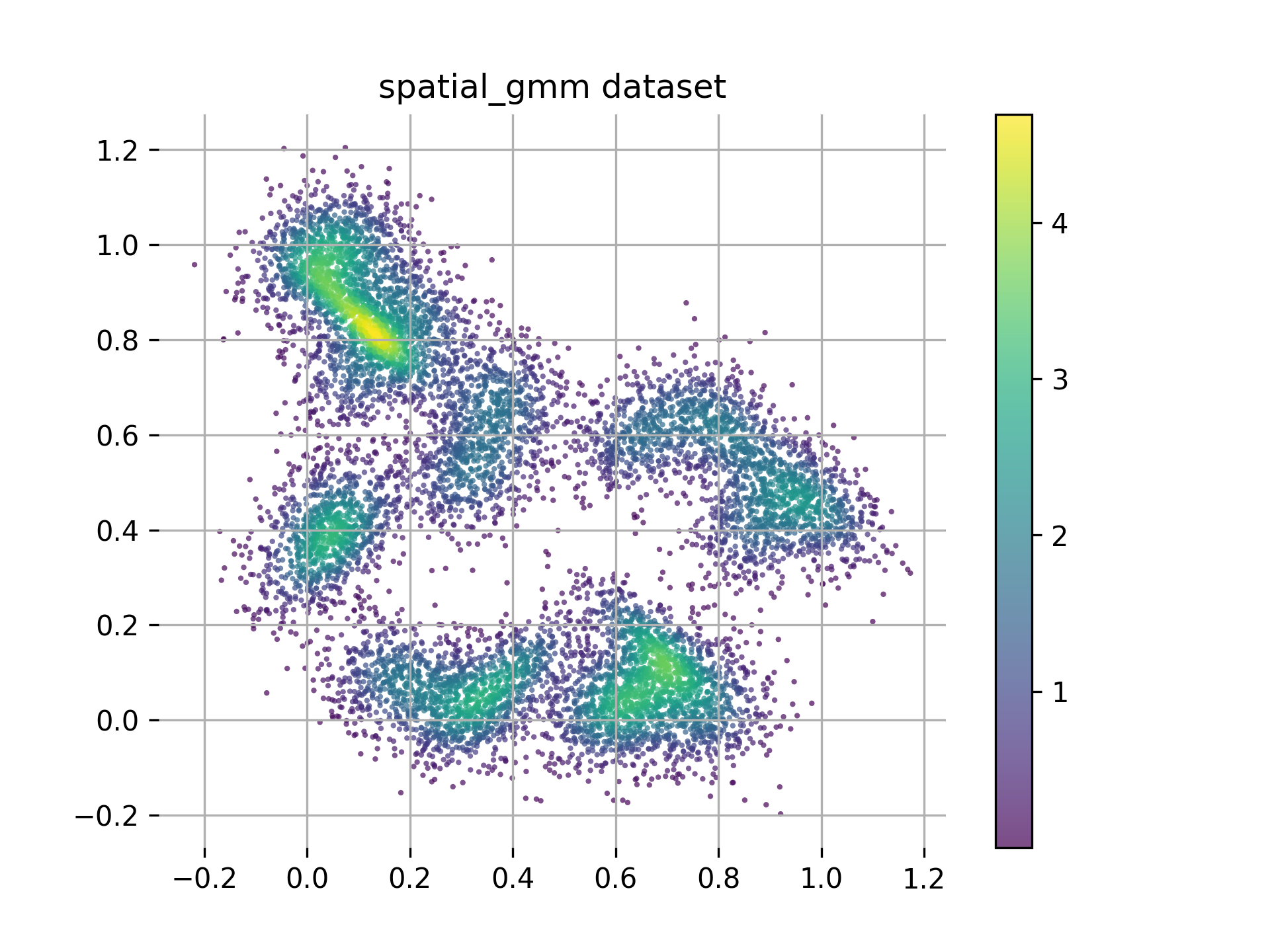}
\includegraphics[scale=0.43]{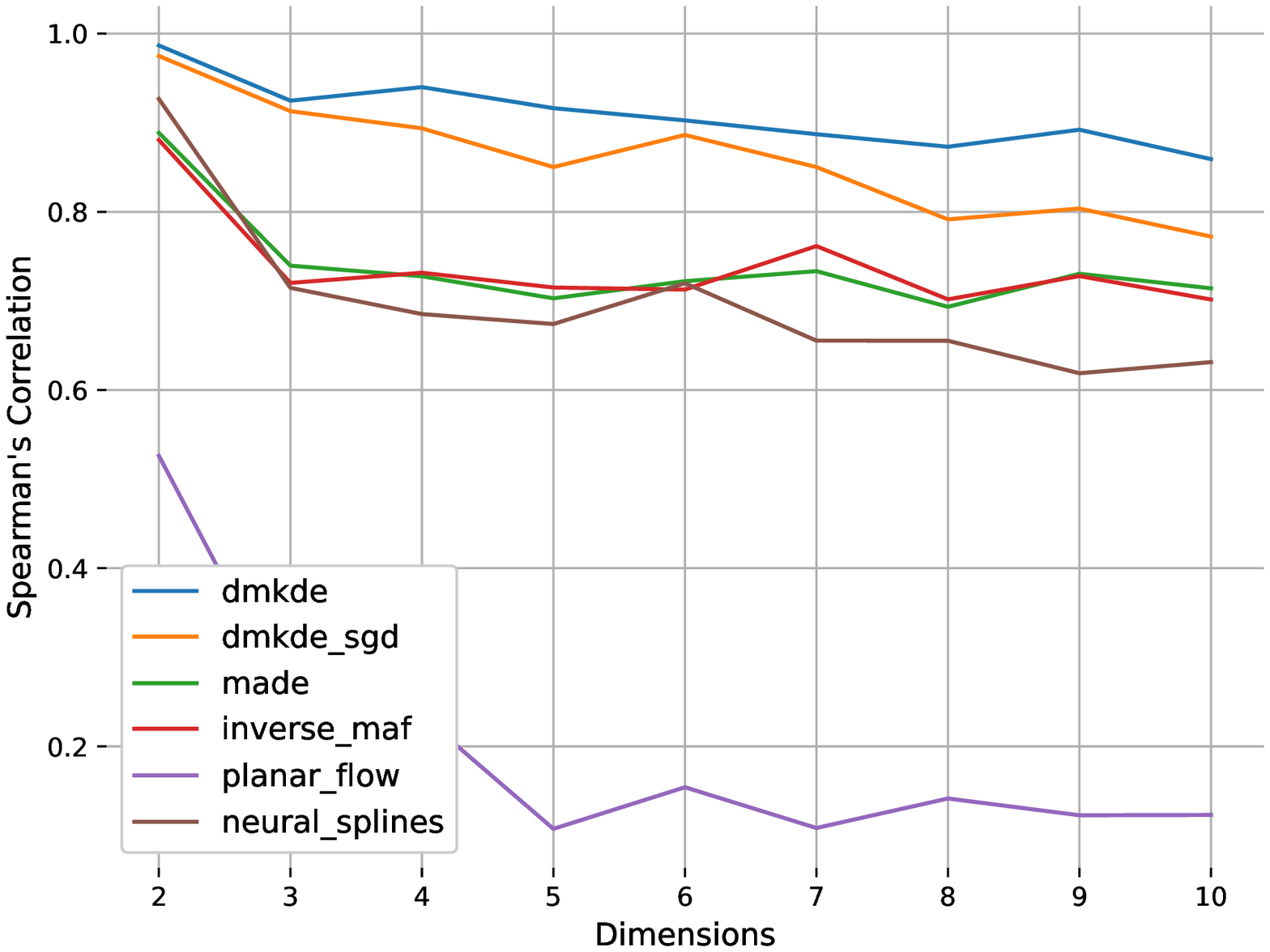}
\par\end{centering}
\caption{(left) Random generated samples, (right) experimental results of unconditional random density estimation in higher dimensions. The x-axis is the dimensions and the y-axis is the Spearman's correlation. \label{fig:spatial-gmm-results}}
\end{figure}

Figure \ref{fig:spatial-gmm-results} shows the results obtained by each algorithm in the random Gaussian mixture model synthetic data set. \textbf{Planar flow} has the worst performance among the six algorithms. \textbf{Made} and \textbf{Inverse Maf} have a similar behavior. They start at nearly 0.92 of Spearman's correlation and decrease with higher dimensions. In ten dimensions obtain nearly 0.76 of Spearman's correlation. \textbf{Neural Splines} starts better than \textbf{Made} and \textbf{Inverse Maf}; however, its performance decrease with more dimension. It shows worst performance in ten dimension compared to \textbf{Made} and \textbf{Inverse Maf} obtaining 0.61 of Sperman's correlation. \textbf{QAFFDE} start with nearly 1.0 Spearman's correlation and drop slightly with more dimensions. However, its performance is the best among all the neural density estimation methods. \textbf{QAFFDE-SGD} has similar behavior than QAFFDE but with lower Spearman's correlation.

\subsection{Conditional Density Estimation} \label{sect:cond_density_estimation}

Density estimation can be used as a conditional density algorithm (conditioned on class values). In this subsection, we present a systematic evaluation of the conditional density estimation obtained by QAFFDE and compared it against state-of-the-art neural flow methods in two frequently used benchmark image data sets MNIST and CIFAR. 

\subsubsection{Data sets and experimental setup}

\begin{table}[tbh]
\caption{Data sets used for conditional density estimation.}
\label{Table:conditional_datasets}
\vskip 0.15in
\begin{center}
\begin{small}
\begin{sc}
\begin{tabular}{lcccr}
\toprule
Data set & Attributes & Classes & Train-Test \\
\midrule
Mnist & 784 & 10 & 60000-10000 \\ 
Cifar & 3072 & 10 & 60000-10000 \\ 
\bottomrule
\end{tabular}
\end{sc}
\end{small}
\end{center}
\vskip -0.1in
\end{table}

Two benchmark image data sets were used. The details of these data sets are shown in Table \ref{Table:conditional_datasets}. QAFFDE was trained using a conditional Bayesian density estimation strategy and ADAM as the stochastic optimization gradient algorithm. As a baseline we compared QAFFDE algorithms against three state-of-the-art normalizing flow algorithms (MADE \citep{germain2015made}, RealNVP \citep{dinh2014nice}, MAF \cite{Papamakarios2017})  and RoundTrip \citep{liu2021density}, which is a normalizing flow generative adversarial algorithm, for further details see Subsection \ref{subsect:neural_flows}. Each algorithm was assessed using the conditional density computed over the test image data set and maximizing the posterior probability conditioning on the class label. Besides, we built a LeNet architecture \citep{lecun1989backpropagation} as a feature extraction method and paired it with QAFFDE-SGD as the density estimation method. The LeNet part had two sequential convolutional layers. Both of them have a kernel size of 5, a same padding and a relu activation function. The first and the second convolutional layer had 20 and 50 filters correspondingly. The third layer is a fully connected layer with a size of 84 neurons. This fully connected layer is joined to the QAFFDE layer. The AFF layer were trained using 1000 adaptive Fourier features. In addition, the AFF layer was only optimized using the algorithm shown in the Section \ref{sec:methods}, but not in the QAFFDE-SGD optimization steps to make all comparisons fair. Thus, all AFF weights were set as untrainable in the QAFFDE-SGD optimization step of the neural network. 

For each data set, we performed a hyper parameter optimization using a cross-validation methodology with 30 randomly generated settings. For tunning the $\gamma$ parameter of QAFFDE, we computed the mean distance between pair of points in every data set and selected an appropriate value for $\gamma=\frac{1}{2\sigma^2}$. The number of adaptive Fourier features was set to 1000 for every QAFFDE algorithm. The learning rate was selected in the interval (0, 0.001]. The number of eigen-components was selected in the list {0, 0.1, 0.5, 1} where each number represents a percentage of the number of AFF. The mean of the accuracy on ten experiments was reported.

\subsubsection{Results and discussion}

\begin{table}[thb]
\caption{Accuracy results in conditional density estimation experiment using neural density estimation methods.}
\label{QMC_results}
\vskip 0.15in
\begin{center}
\begin{small}
\begin{sc}
    \begin{tabular}{lll}
\toprule
\toprule
        Algorithm & MNIST & CIFAR-10 \\ 
        \midrule
        MADE & 0.911 & 0.358 \\ 
        RealNVP & 0.744 & 0.309 \\ 
        MAF & 0.926 & 0.295 \\ 
        RoundTrip (CNN) & 0.983 & 0.427 \\ \hline
        QAFFDE & 0.811 & 0.271 \\ 
        QAFFDE-SGD & 0.952 & 0.484 \\
        LENET QAFFDE-SGD & $\bm{0.989}$ & $\bm{0.628}$ \\ 
        \bottomrule
    \end{tabular}
\end{sc}
\end{small}
\end{center}
\vskip -0.1in
\end{table}

Table \ref{QMC_results} shows the results obtained by each algorithm in the density estimation task for MNIST and CIFAR-10 image datasets. It can be seen that without convolutional neural networks QAFFDE-SGD is better than all other neural flow methods except RoundTrip on Mnist, and it is the best on CIFAR-10. When we use convolutional layers with QAFFDE-SGD its performance outperforms the other methods on both image datasets. The method can be used not only as a density estimation algorithm, but also as a conditional density estimation method. QAFFDE was sistem

\section{Conclusion}
\label{sec:conclusions}

This paper presented a method for density estimation and its systematic evaluation on different density estimation tasks. The method combines two, in principle, different approaches for density estimation. First, it can be seen as an approximated version of KDE that do not require to store all the training samples; and, second, it can be seen as a form of neural density estimation that can be trained using gradient descent and can be integrated with other deep learning architectures. The method combines two ideas from seemingly unrelated fields, density matrices used in quantum mechanics to represent the state of a quantum system, and random features a method to efficiently approximate kernels in machine learning. QAFFDE was systematically evaluated and compared against state-of-the-art neural density estimation methods, on four different experiments. QAFFDE showed a competitive performance through all the experiments, performing on par and in some cases outperforming state-of-the-art methods.

% Acknowledgements should only appear in the accepted version.
\section*{Acknowledgements}

\bibliography{article}
\bibliographystyle{icml2021}

\end{document}